\RequirePackage{fix-cm}
\documentclass[smallextended]{svjour3}       
\smartqed  
\usepackage{graphicx}
\begin{document}

\title{ Semantic Knowledge Discovery and Discussion Mining of Incel Online Community: Topic modeling 
}
\subtitle{ \\  }


\author{ Hamed Jelodar, Richard Frank
}


\date{Received: date / Accepted: date}

\institute{H. Jelodar \at
              School of Computer Science and Technology, Nanjing University of Science and Technology, Nanjing 210094, China \\
              \email{jelodar@njust.edu.cn}           
           \and
          R. Frank \at
              School of Criminology, Simon Fraser University, Burnaby, British Columbia, Canada\\
               \email{rfrank@sfu.ca}
}

\date{Received: date / Accepted: date}

\maketitle

\begin{abstract}
Online forums provide a unique opportunity for online users to share comments and exchange information on a particular topic. Understanding user behaviour is valuable to organizations and has applications for social and security strategies, for instance, identifying user opinions within a community or predicting future behaviour. Discovering the semantic aspects in Incel forums are the main goal of this research; we apply Natural language processing techniques based on topic modeling to latent topic discovery and opinion mining of users from a popular online Incel discussion forum. To prepare the input data for our study, we extracted the comments from Incels.co. The research experiments show that Artificial Intelligence (AI) based on NLP models can be effective for semantic and emotion knowledge discovery and retrieval of useful information from the Incel community. For example, we discovered semantic-related words that describe issues within a large volume of Incel comments, which is difficult with manual methods.

\keywords{Natural Language Processing \and Incel \and LDA \and Social Media \and Online Forums}
\end{abstract}

\section{Introduction}
Online forums and social media platforms are becoming a gold source of comments and the opinions of people. However, it would be a great task to be able to extract narratives based on systematic summarization that can help the retrieval of useful information from online communities [1-5]. In this research, we focused on semantic aspects of the online incel community. However, incels are members of an self-identified exclusive online culture, because they are unable to find sexual relationships and/or have problems in engaging with romantic partners.  The word 'incel' is created from the terms 'involuntary celibate'. The phenomenon became known to  people in the social community  in Canada and America after the mass murders in 2018 [6] [7] [8]. Moreover, the incel culture is characterized by sexual frustration, racism, anti-feminism, misogynistic violent extremism and they are also closely linked to the subject of online hate culture [9]. In this paper, our intention is to investigate the incel culture using artificial intelligence (AI) methods such as topic modeling to knowledge discovery to better understand the social challenges related to incel discussions in online communities.\\\\

\begin{table}[]
\center
\caption{Example of incel comments.}
\begin{tabular}{|l|l|}
\hline
User ID & \multicolumn{1}{c|}{incel Comment}                                                                                                                                                                                        \\ \hline
50320   & \begin{tabular}[c]{@{}l@{}}We know they hate ugly males but do you think they hate ugly \\ females as well? Or is it just with men?\end{tabular}                                                                          \\ \hline
50932   & \begin{tabular}[c]{@{}l@{}}Im fucking dying on the inside. I fucking hate walking past the\\ mirrors  and seeing my ugly fucking face. I need to die.\end{tabular}                                                        \\ \hline
57023   & \begin{tabular}[c]{@{}l@{}}I understand why women hate ugly men. But why do men hate us, \\ too? Why are they programmed like that? There \\ has to be a biological/scientific reason behind that behaviour.\end{tabular} \\ \hline
59087   & \begin{tabular}[c]{@{}l@{}}No doubt. They hide behind the "misogynist, sexist" as an \\ acceptable excuse.\end{tabular}                                                                                                   \\ \hline
\end{tabular}
\end{table}

Definitely, AI based on natural language processing (NLP) techniques can have a practical application for understanding the semantic and emotional aspects, and also for the retrieval of useful information from within the incel comments [10-12]. We can take advantage of the semantic models through i) topic modelling of short-text topics, ii) analysis of incel comments, iii) semantic knowledge discovery and iv) extracting meaningful topics. Recently, online hate speech within incel communities has risen to become a new challenge for cyber-crime research. In this research, we present a systematic model to provide insight into the technologies based on NLP models to reduce these challenges. Moreover, we provide a new dataset of 18,097 unique comments containing incel-related texts of a famous online forum (for examples, see Table 1). The main contributions of this paper are summarized as follows:
\begin{itemize}

\item We show a novel application of AI models to discover the meaningful latent topics drawn from comments from a large online incel community.
\item A new incel dataset is presented in this research, which can help future researchers perform further analysis and work on this area. 
\item In this context, we discover the semantic relationships between Incel discussions and latent topics to more understand this online community.

\end{itemize}
\section{Background and Related works}
In this section, we first provide a background on the application of AI and machine learning models on the incel comments. Then, we present an overview of NLP methods and text summarization.

\subsection{Incels and Their Ideology}
An 'incel' is defined as a person who has a continuing problem of finding a romantic or sexual partner, despite desiring one. The word is used to define an extreme faction of the Men’s Rights Movement and its membership, which is intertwined with shared experiences of loneliness and sexlessness [13] [14]. In 2018, a massacre in Toronto, Canada killed nine people, and, after police investigation, the person claimed that he had been radicalized online and subscribed to the incel ideology. This is an example of how the underlying concepts of the incel ideology may have dire consequences in the real world [15] [16].\\

\subsection{The Incel Community and the Web}
Recently some authors focused on YouTube to analyse the incel community by evaluating this community over the last decade and understanding of YouTube recommendation algorithm direct online users about  incel-related videos [16]. To understand this ideology on YouTube, they gathered thousands of  videos shared by online users from Reddit. Finally, found a non-negligible growth in incel-related content on YouTube over the past few years [16]. In another research the authors evaluated the people comments from incels.me to investigate the  incel movement nature [17].\\

In addition, in [18] the authors introduced a dataset to detect involuntary celibates (incels) in social networks or identify people who think women are not attracted to them in social networks. Also, other researchers focused on 8,000 incel posts to find the values, and beliefs of this community with key terms related to incel topics and using Google search engine [19].

\section{Research Model and Question}
The main motivation of this paper is to reduce the recent challenges in incel research by investigating differences in misogyny and the role of incel forums in radicalization, which can be effective to better understand the extremism aspects and online subculture of the incel ideology [20-22]. However, by understanding the types of characters and retrieving meaningful topics from these communities, one might be able to design social strategies based on AI-models to prevent their hostility towards women. The framework presented in this paper is a data-based system including the collection of incel-comments, data pre-processing, NLP analytics and data visualisation to generate insight related to incel-comments. However, this framework is implemented to answer the following research questions (RQ):
\begin{itemize}

\item RQ 1: How can topic modeling AI models be used for finding the highlighted issues and discovering semantic-topics within active Incel comments? 

\item RQ 2: What is the semantic-topic distribution of incel comments on various time, from this popular online forum ? 

\end{itemize}

\subsection{Collecting and Extracting} 
In the first step of this research, we collected our data from incels.co, an online incel forum, which is a support website for people who are unable to have sex and romantic relationships, or  most of them rely on people who are not their girlfriend/not married. The dataset includes all the responses, posts (comments), along with the user's profiles who completed the official registration on this website.

\subsection{Data Processing \& Text Normalization}
 Text Normalization is one of the important steps to prepare data and text processing in natural language processing. It is an actuality that unstructured content in incel comments include big amounts of noise words. incel users are susceptible to typographical errors, make typing mistakes, use shortenings (of a word) or slang, and use punctuation signs to show emotions (such as exclamation marks). Usually, it is not essential to have all of the words in the topic modeling step, and some can be replaced or ignored. 

\subsection{Removing Noise and Input Pre-processing}
There are various methods to clean noise from text documents. To get over this noise, we considered stop-words, Stemming methods and also used Natural Language Toolkit (NLTK) to perform these tasks. However, after sentence segmentation, all the stop-words have been eliminated from the original text-documents. Based on our goal of this research, the stop words are listed as words which generally refer to the most frequent words in a language. According to our knowledge, there is no standard list of stop words as stop words in different languages are special and unique to that language.

\subsection{ NLP and identifying the highlighted Incel issues} 
Topic modeling is a powerful technique in NLP and text processing to analyze huge volumes of text contents. Topic modeling methods aim to decrease the size of the feature space by clustering words of a text-corpus to a relatively few topics, and then representing each document as a mixture of other topics [23-27]. In this step, we use Latent Dirichlet Allocation (LDA) [28], which is most well-known for semantic mining and latent topic discovery in natural language processing. Based on LDA, we define a set of documents (where in our case each document is a post from the incel forum) and words as topics. For LDA, given a vocabulary with $N$ words, $D=w_{1} ,w_{2} ,.....,w_{N} $, that each Incel-comment $w$ in a corpus $D$. Given the parameters$\; \alpha \;$and$\; \beta $, and the joint distribution over the random variables $(w_{m} ,z_{m} ,\varphi _{k} ,\theta _{m} )$, which is given by:
\begin{equation} \label{GrindEQ__2_}
\begin{array}{l} {p\left(w_{m} ,z_{m} ,\theta _{m} ,\varphi _{k} |\alpha ,\beta \right)=p\left(\theta _{m} |\alpha \right)} \\ {p(\varphi _{k} |\beta )\mathop{\mathop{\mathop{\prod  }\limits_{} }\limits^{N_{m} } }\limits_{n-1} {\kern 1pt} p(z_{m,n} |\theta _{m} )p(w_{m.n} |\varphi _{z_{m,n} } ),} \end{array}
\end{equation}

Embedding over $\theta _{m} \;$and$\; \varphi _{k} $, summing over $z_{m,n} $ , the marginal distribution of an incel-comments. Overall, LDA is a generative probabilistic approach where each text document (incel comments) is represented as a finite mixture over a collection of (latent) topics and each topic is explained by a distribution on words. 

\subsubsection{Incel Topic Distribution of Various Years}
As an important issue in topic distribution, we need to provide information about the distribution of documents over topics, and topics over words. To solve this issue, we take advantage of the Gibbs sampling algorithm [29] on our topic modeling. This method is a Monte Carlo Markov-chain algorithm, for generating a sample from a joint distribution, and each parameter is sequentially sampled according on the viewed data and all the other parameters.

\subsubsection{Incel-Topic Labelling}
Labeling the incel-topics is essential for obtaining insights after extracting the most significant terms. As an example of use, by labeling the topics one could create a clear insight that can understand the whole corpus by investigating the generated topics. However, LDA does not tend to have a direct label for the topics because each topic is described as a probability distribution. Since, human estimation is one of the most trusted approaches for labeling and determining the quality of generated topics, thus, the labels were assigned manually, by inspecting and considering the top relevant words. We also recruited 4 human domain experts to validate the label of each topic. 

\section{Experiments and Results}
In order to obtain the main research goal of this study, we combined semantic and behaviour analysis of the incel comments from a famous online forum, incels.co. We used a Web intelligence platform, housed at the International CyberCrime Research Centre (ICCRC) at Simon Fraser University, to extract the incel-comments from the online forum. Data was collected between November, 2017 and March, 2020 with a total of 18,097 unique comments. To implement the LDA model, we used Mallet Package with 1000 iterations, an alpha value of 0.05 and a beta value of 0.01.  By considering various experiments, the best results were achieved via 100 topics and 20 terms per topic. 

\subsection{Generating Meaningful Labels and Incel-topic Distribution}
Regarding the research questions, we discovered the most frequently mentioned topics, which were mentioned by users of the online forum. The topic distribution can be viewed as a latent representation of the comments by the incels, which can be used as a feature for prediction, such as semantic/emotion analysis. Therefore, in this part of the research, we describe the informative and statistic results to present the findings of the paper.

\begin{figure}
\centering

\includegraphics[ height=9cm, width=12cm]{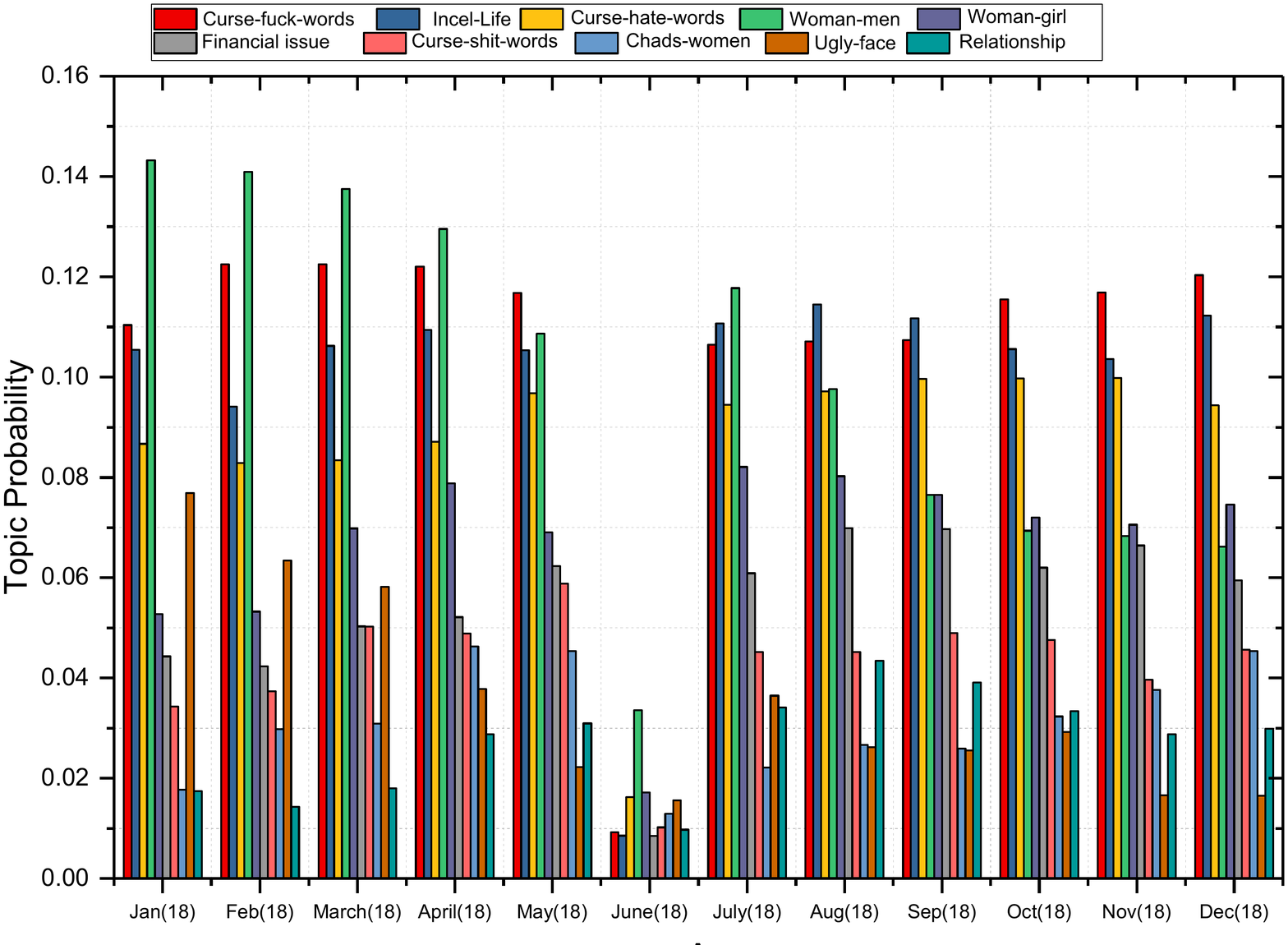}
\caption{Time trends of meaningful-topics relted Incel topics in 2018}
\label{fig:1}       
\end{figure}

\begin{figure}

\centering

\includegraphics[ height=9cm, width=12cm]{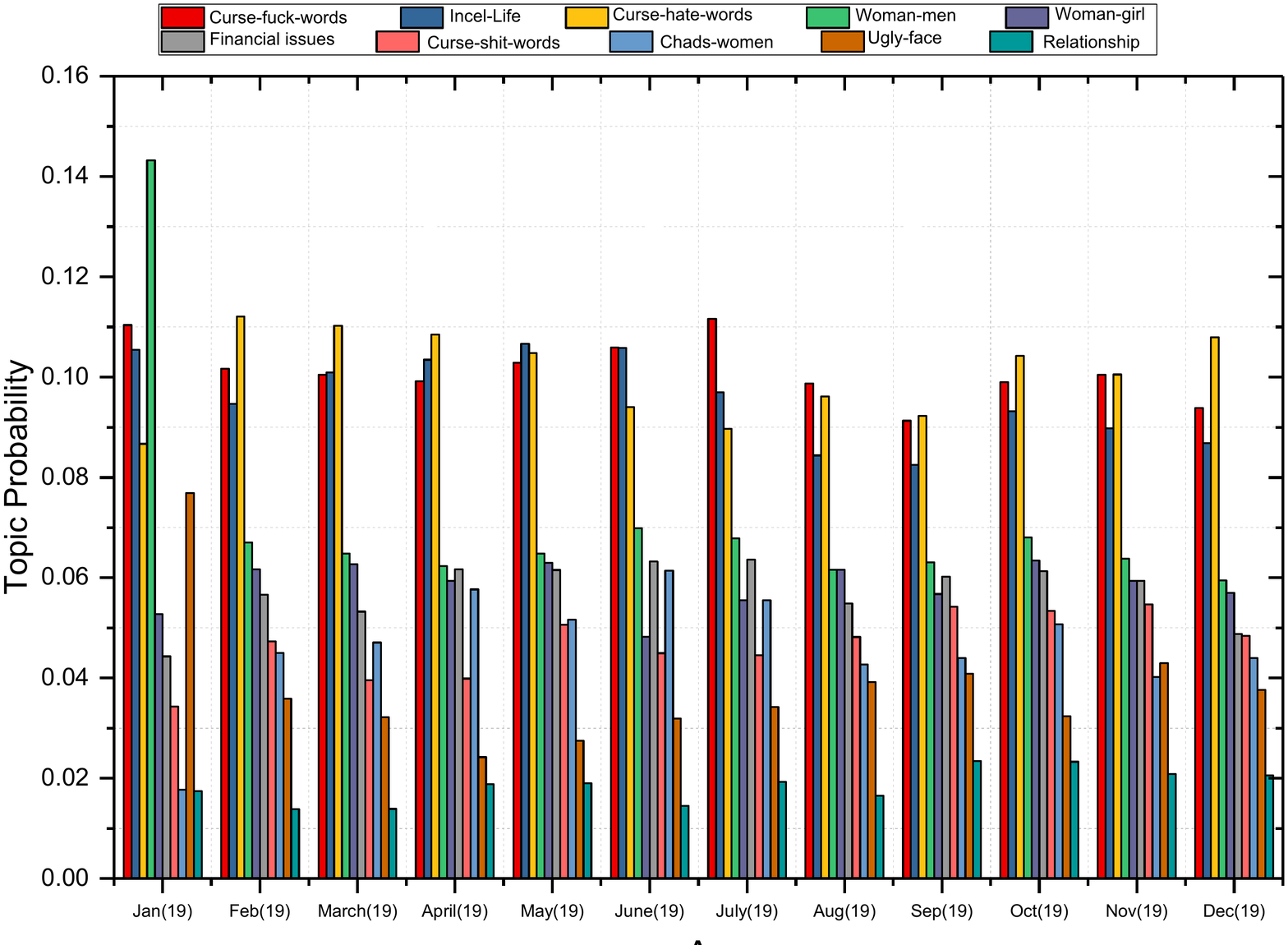}

\caption{Time trends of meaningful-topics meaningful-topics relted Incel topics in 2019}
\label{fig:1}       
\end{figure}

\begin{table}
\centering
\caption{Topic discovered of the from the ranks 1-10 ( TW: Topic Words , PTW: Prior topic weights).}

\resizebox{12cm}{!} {
\begin{tabular}{|c|c|c|c|c|c|l|l|l|l|} 
\hline
\multicolumn{2}{|l|}{~Topic 30}                                                                                                                                                                                                                                                               & \multicolumn{2}{l|}{~Topic 83}                                                                                                                                                                                                                                                                & \multicolumn{2}{l|}{~Topic 15}                                                                                                                                                                                                                                                   & \multicolumn{2}{l|}{~Topic 54}                                                                                                                                                                                                                   & \multicolumn{2}{l|}{~Topic 7}                                                                                                                                                                                                                                          \\ 
\hline
\multicolumn{2}{|c|}{PTW (3.73729}                                                                                                                                                                                                                                            & \multicolumn{2}{c|}{PTW (3.5824)}                                                                                                                                                                                                                                             & \multicolumn{2}{c|}{PTW (2.7915)}                                                                                                                                                                                                                                & \multicolumn{2}{l|}{PTW (2.50559)}                                                                                                                                                                                               & \multicolumn{2}{l|}{PTW (2.26295)}                                                                                                                                                                                                                     \\ 
\hline
TW                                                                                                                                                & TW                                                                                                                                        & TW                                                                                                                                             & TW                                                                                                                                           & TW                                                                                                                                        & TW                                                                                                                                   & TW                                                                                                                   & TW                                                                                                                        & TW                                                                                                                       & TW                                                                                                                                          \\ 
\hline
\begin{tabular}[c]{@{}c@{}} good \\Incel\\~bad \\fuck \\shit \\pretty \\long \\real \\face \\guy \end{tabular}                                    & \begin{tabular}[c]{@{}c@{}}place \\great \\care \\guys \\fucking \\low \\kind \\true \\person\\~hate~ ~ ~ ~~\end{tabular}                 & \begin{tabular}[c]{@{}c@{}} women \\ugly\\~Incel \\fuck \\fucking \\sex \\shit \\guy \\life \\men \end{tabular}                                & \begin{tabular}[c]{@{}c@{}}woman \\face \\cuck \\guys \\normie \\average \\normies \\blackpill \\cope \\fat~ ~ ~ ~ ~ ~\end{tabular}          & \begin{tabular}[c]{@{}c@{}} life\\~school \\chad \\good \\shit \\girls \\love \\fucking \\friends \\years ~\end{tabular}                  & \begin{tabular}[c]{@{}c@{}}parents \\girl \\live \\worse \\friend \\chads \\told \\hate \\kids \\bad\end{tabular}                    & \begin{tabular}[c]{@{}l@{}} good\\~Incels \\money \\cope \\years \\buy \\long \\home \\nice \\night \end{tabular}    & \begin{tabular}[c]{@{}l@{}}started \\hours \\times \\sleep \\stuff \\months \\hair \\live \\sounds \\half~\end{tabular}   & \begin{tabular}[c]{@{}l@{}} iq \\life\\~cope \\thread \\live \\ugly \\death \\mind \\kill \\low ~\end{tabular}           & \begin{tabular}[c]{@{}l@{}}autism \\aren \\change \\person \\society \\die \\autistic \\average \\free \\sense\end{tabular}                 \\ 
\hline
\multicolumn{2}{|c|}{Topic 67~}                                                                                                                                                                                                                                                               & \multicolumn{2}{c|}{~Topic 23}                                                                                                                                                                                                                                                                & \multicolumn{2}{c|}{Topic 42~~}                                                                                                                                                                                                                                                  & \multicolumn{2}{l|}{~ Topic 38}                                                                                                                                                                                                                  & \multicolumn{2}{l|}{~ Topic 26}                                                                                                                                                                                                                                        \\ 
\hline
\multicolumn{2}{|l|}{PTW (2.10407)}                                                                                                                                                                                                                                           & \multicolumn{2}{l|}{PTW (1.72444)}                                                                                                                                                                                                                                            & \multicolumn{2}{l|}{PTW (1.51489}                                                                                                                                                                                                                                & \multicolumn{2}{l|}{PTW (0.94097)}                                                                                                                                                                                               & \multicolumn{2}{l|}{PTW (0.93482)}                                                                                                                                                                                                                     \\ 
\hline
\multicolumn{1}{|l|}{TW}                                                                                                                          & \multicolumn{1}{l|}{TW}                                                                                                                   & \multicolumn{1}{l|}{TW}                                                                                                                        & \multicolumn{1}{l|}{TW}                                                                                                                      & \multicolumn{1}{l|}{TW}                                                                                                                   & \multicolumn{1}{l|}{TW}                                                                                                              & TW                                                                                                                   & TW                                                                                                                        & TW                                                                                                                       & TW                                                                                                                                          \\ 
\hline
\multicolumn{1}{|l|}{\begin{tabular}[c]{@{}l@{}} women \\men\\~chad \\Incels \\male \\woman \\chads \\attractive \\love \\females ~\end{tabular}} & \multicolumn{1}{l|}{\begin{tabular}[c]{@{}l@{}}female \\guys \\good \\girl \\girls \\males \\sexual \\sex \\society \\years\end{tabular}} & \multicolumn{1}{l|}{\begin{tabular}[c]{@{}l@{}} social\\~Incels \\men \\friends \\normies \\age \\low \\society \\young \\group \end{tabular}} & \multicolumn{1}{l|}{\begin{tabular}[c]{@{}l@{}}years \\big \\media \\inhib \\normie \\school \\play \\experience \\gym \\dude~\end{tabular}} & \multicolumn{1}{l|}{\begin{tabular}[c]{@{}l@{}}~fucking\\~fuck \\kill \\hate \\death \\shit \\life \\die \\female \\women ~\end{tabular}} & \multicolumn{1}{l|}{\begin{tabular}[c]{@{}l@{}}ass \\dick \\love \\dog \\cucks \\live \\hope \\water \\real \\children\end{tabular}} & \begin{tabular}[c]{@{}l@{}} foids \\foid\\~chad \\men \\chads \\fuck \\cucks \\iq \\cucked \\whores ~ ~\end{tabular} & \begin{tabular}[c]{@{}l@{}}life \\imagine \\mogs \\blackpilled \\angry \\ascend \\west \\normies \\smv \\mog\end{tabular} & \begin{tabular}[c]{@{}l@{}} sex \\social\\human \\skin \\child \\culture \\sexual \\power \\virgin \\god ~~\end{tabular} & \begin{tabular}[c]{@{}l@{}}mental \\science \\level \\experience \\extremely \\hate \\wanted \\moral \\reality \\relationship\end{tabular}  \\
\hline
\end{tabular}
}
\end{table}

\begin{table}
\centering
\caption{Topic discovered from the ranks 11-20 ( TW: Topic Words , PTW: Prior topic weights).}
\resizebox{12cm}{!} {
\begin{tabular}{|c|c|c|c|c|c|l|l|l|l|} 
\hline
\multicolumn{2}{|l|}{~Topic~18}                                                                                                                                                                                                                                                & \multicolumn{2}{l|}{~Topic 55~}                                                                                                                                                                                                                                              & \multicolumn{2}{l|}{~Topic 56~}                                                                                                                                                                                                                                                                           & \multicolumn{2}{l|}{~Topic 72}                                                                                                                                                                                                                                & \multicolumn{2}{l|}{Topic 99~}                                                                                                                                                                                                                         \\ 
\hline
\multicolumn{2}{|c|}{PTW (0.9251)}                                                                                                                                                                                                                             & \multicolumn{2}{c|}{PTW (0.9112)}                                                                                                                                                                                                                            & \multicolumn{2}{c|}{PTW (0.88379)}                                                                                                                                                                                                                                                        & \multicolumn{2}{l|}{PTW (0.80747)}                                                                                                                                                                                                            & \multicolumn{2}{l|}{PTW (0.75448)}                                                                                                                                                                                                     \\ 
\hline
TW                                                                                                                                    & TW                                                                                                                                     & TW                                                                                                                                    & TW                                                                                                                                   & TW                                                                                                                                               & TW                                                                                                                                                     & TW                                                                                                                              & TW                                                                                                                          & TW                                                                                                                           & TW                                                                                                                      \\ 
\hline
\begin{tabular}[c]{@{}c@{}}girls~\\chad~\\~girl~\\women~\\Incel~\\guys~\\guy~\\fucking~\\fuck~\\femoids~~\end{tabular}                & \begin{tabular}[c]{@{}c@{}}sex~\\chads~\\date~\\laid~\\legit~\\game~\\femoid~\\friends~\\attractive~\\fakecel\end{tabular}             & \begin{tabular}[c]{@{}c@{}} fuck\\~fucking \\shit \\thread \\gay \\lmao \\black \\ass \\guy \\cuck ~\end{tabular}                     & \begin{tabular}[c]{@{}c@{}}video \\weed \\kek \\guys \\big \\dude \\woman \\rape \\threads \\bitch\end{tabular}                      & \begin{tabular}[c]{@{}c@{}} money \\job\\~life \\shit \\neet \\pay \\years \\society \\live \\jobs ~\end{tabular}                                & \begin{tabular}[c]{@{}c@{}}living \\rich \\free \\government \\family \\cucks \\good \\market \\paying \\low\end{tabular}                              & \begin{tabular}[c]{@{}l@{}} black\\~reddit \\gun \\videos \\internet \\youtube \\online \\happened \\easy \\chad ~\end{tabular} & \begin{tabular}[c]{@{}l@{}}case \\guns \\video \\fight \\watching \\body \\america \\eyes \\site \\google\end{tabular}      & \begin{tabular}[c]{@{}l@{}} black\\~whites \\ethnic \\women \\asian \\curry \\ethnics \\curries \\blacks \\iq ~\end{tabular} & \begin{tabular}[c]{@{}l@{}}cope \\skin \\asians \\jbw \\west \\low \\girls \\indian \\fucking \\guys\\\end{tabular}     \\ 
\hline
\multicolumn{2}{|c|}{Topic~65~}                                                                                                                                                                                                                                                & \multicolumn{2}{c|}{Topic 17~}                                                                                                                                                                                                                                               & \multicolumn{2}{c|}{Topic 39~}                                                                                                                                                                                                                                                                            & \multicolumn{2}{l|}{~Topic~98~}                                                                                                                                                                                                                               & \multicolumn{2}{l|}{~Topic 4}                                                                                                                                                                                                                          \\ 
\hline
\multicolumn{2}{|l|}{PTW (0.69627)}                                                                                                                                                                                                                            & \multicolumn{2}{l|}{PTW (0.55989)}                                                                                                                                                                                                                           & \multicolumn{2}{l|}{PTW (0.54066)}                                                                                                                                                                                                                                                        & \multicolumn{2}{l|}{PTW (0.4959)}                                                                                                                                                                                                             & \multicolumn{2}{l|}{PTW (0.47992)}                                                                                                                                                                                                     \\ 
\hline
\multicolumn{1}{|l|}{TW}                                                                                                              & \multicolumn{1}{l|}{TW}                                                                                                                & \multicolumn{1}{l|}{TW}                                                                                                               & \multicolumn{1}{l|}{TW}                                                                                                              & \multicolumn{1}{l|}{TW}                                                                                                                          & \multicolumn{1}{l|}{TW}                                                                                                                                & TW                                                                                                                              & TW                                                                                                                          & TW                                                                                                                           & TW                                                                                                                      \\ 
\hline
\multicolumn{1}{|l|}{\begin{tabular}[c]{@{}l@{}} face \\good\\~ugly \\cope \\eyes \\surgery \\chad \\nose \\eye \\chin \end{tabular}} & \multicolumn{1}{l|}{\begin{tabular}[c]{@{}l@{}}jaw \\pretty \\hair \\guy \\facial \\money \\bad \\average \\fuck \\age~~\end{tabular}} & \multicolumn{1}{l|}{\begin{tabular}[c]{@{}l@{}} shit \\Incel\\~good \\forum \\avi \\gay \\thread \\fuck \\iq \\banned ~\end{tabular}} & \multicolumn{1}{l|}{\begin{tabular}[c]{@{}l@{}}free \\nice \\low \\Incels \\mogs \\virgin \\soy \\cool \\fun \\posting\end{tabular}} & \multicolumn{1}{l|}{\begin{tabular}[c]{@{}l@{}} jews\\~jewish \\countries \\europe \\hitler \\jew \\west \\hate \\whites \\israel \end{tabular}} & \multicolumn{1}{l|}{\begin{tabular}[c]{@{}l@{}}cucked \\left \\western \\germany \\history \\feminism \\alt \\german \\nazi \\political~\end{tabular}} & \begin{tabular}[c]{@{}l@{}} good \\anime\\~game \\play \\shit \\indari \\music \\playing \\love \\played ~\end{tabular}         & \begin{tabular}[c]{@{}l@{}}watching \\school \\life \\video \\pretty \\hate \\normie \\movie \\great \\normies\end{tabular} & \begin{tabular}[c]{@{}l@{}} age \\girls\\~sex \\young \\children \\girl \\child \\years \\consent \\older \end{tabular}      & \begin{tabular}[c]{@{}l@{}}kids \\rape \\attracted \\foids \\teen \\yo \\olds \\adult \\younger \\sexual~\end{tabular}  \\
\hline
\end{tabular}
}
\end{table}

Regarding the main goals of this research, we can see the semantic related-words about incel topics (shown in Tables 2-3) which only considered the top 10 topics for each year. Following these results we have:

\begin{figure}

\centering

\includegraphics[ height=6cm, width=9cm]{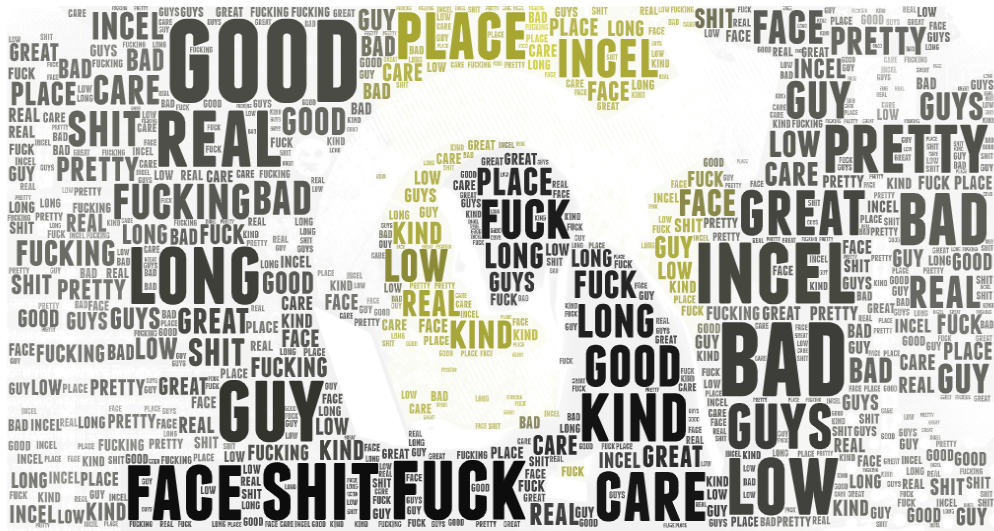}
\includegraphics[ height=6cm, width=9cm]{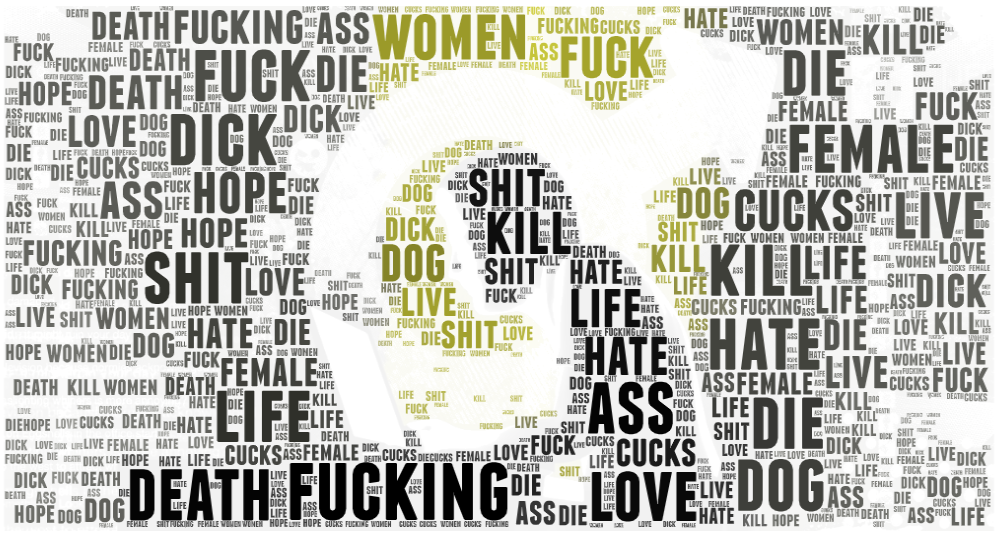}
\includegraphics[ height=6cm, width=9cm]{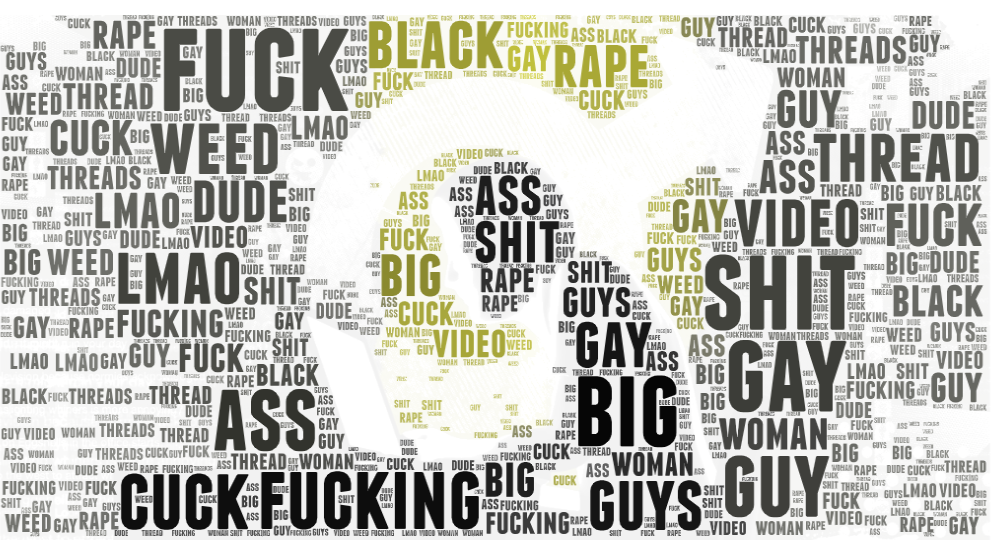}

\caption{Word-topic clouds based on weight of words. From up-side to down-side; Topic 30, Topic 42 and Topic 55 are respectively.}
\label{fig:1}       
\end{figure}

\textbf{Curse words (Topic 30, Topic 42, Topic 55)}: Based on these topics, it seems that there is a relationship between gender and swearing in the incel community, as showed in Figure 3. The results support our observations that the Incel users tend to use more swear words and more negative adjectives such as \textit{shit}, \textit{fuck}, \textit{hate}. However, the swear words are aimed towards, and directly spoken to, other people to insult or offend them (girls/women). For example in topic 42, the swear words appeared in bold type in the dialogues and included: \textit{fucking}, \textit{fuck}, \textit{kill}, \textit{hate}, \textit{death}, \textit{shit}, \textit{life}, \textit{die}, \textit{female}, \textit{women}, \textit{ass}, \textit{dick}, and \textit{love}. However, swearing in social media can be linked to an abusive issues, when it is intended to harm, intimidate or cause emotional or psychological hurt [30].\\

\textbf{Relationships and romance issues (Topic 83, Topic 26)}: Love or romance issues is one of highlighted topics among Incel discussions. However, many Incels desperately want romance. And, most of them are suffering healthy relationship issues [31].\\

\textbf{ Financial issues (Topic 56)}: some terms are reflect financial issues, related to money, job, and payment,and and seems that many Incels have discussions with discuss negatively (have negative-sentiment) about losing their jobs, payment problems, which can be a concern once incels share their experience in the online forums. For example in topic 56, the financial words appeared in bold type in the dialogues and included: \textit{money}, \textit{job}, \textit{life}, \textit{shit}, \textit{pay}, \textit{society}, \textit{live},\textit{ jobs}, \textit{living}, \textit{rich}.\\

\textbf{Relationship and human sexuality (Topic 67, Topic 4, Topic 26)}: One of the main problems of incels who cannot find sex, or a girlfriend, and get sex, is often a side note to the real problem of incels [32].\\

\begin{figure}

\centering

\includegraphics[ height=6cm, width=9cm]{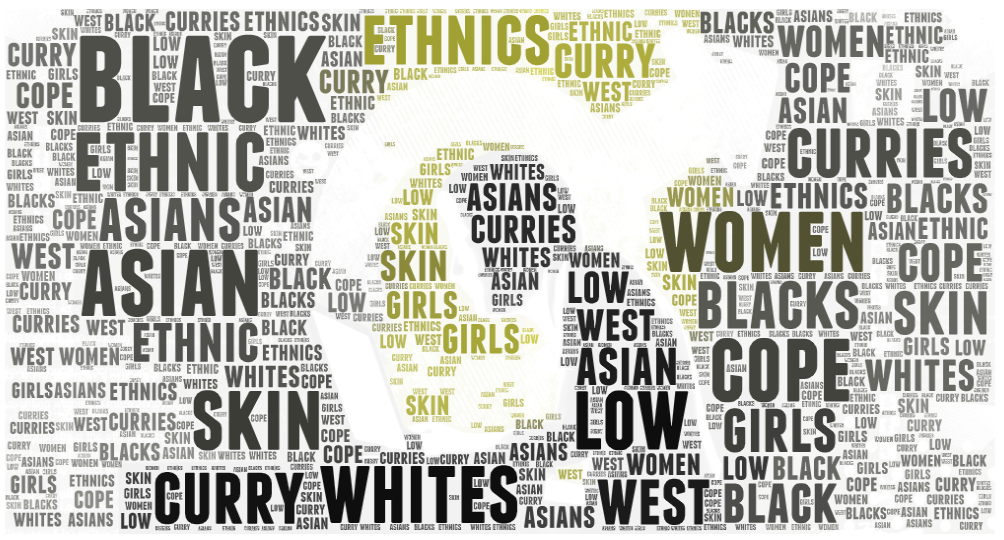}

\caption{Word-topic clouds based on weight of words of Topic 99.}
\label{fig:1}       
\end{figure}

\textbf{Racism issues (Topic 99, Topic 39)}: Regarding theses topics,  it seems that a part of incel discussions are focused on racist-hate issues based on skin-color and ethnic aspects. For example, Topic 99 marked by top terms such as \textit{black}, \textit{whites}, \textit{ethnic}, \textit{women}, \textit{Asian}, \textit{curry}, as showed in Figure 4. However, it is not easy for providing a clear description of racism issues. Based on [34-35], Incels  occupy  a  noteworthy position  within the  proliferation  of  groups  within  the  manosphere,  as  their  expressions  of anti-feminism/racism are among the most hate-filled.

\section{Discussions and Findings}
However, the NLP models based on semantic techniques that allows to discover semantic knowledge and translate social interpretations among online users in social media [36-38]. This paper aimed at identifying the key characters and interests that are mobilized  of Incel online community. Based on our results we discuss some recommendations and highlight the implications of our findings that can be helpful for researchers in this area and social/medical organizations to help the community.

\subsection{First Research Question} 
To answer the RQ1, we generate semantic topics  related to Incel-comments from a popular forum, which implanted a topic modeling based on LDA model for this procedure. Because, LDA has been successfully applied in several applications such as topic discovery, temporal semantic trends, document classification, and finding relations between documents in online community. In this research, the 100 top topic related with each meaningful-topic based on Incel discussions were investigated to understand the means for semantic related-words. To illustrate the objectives, a selection of the top 20 topics that are explicitly related to Incel's issues is shown in Table 2-3.

\subsection{Second Research Question}
For investigating the RQ2, we consider another advantage of NLP methods based on topic modeling and gibbs sampling, which is the identification of semantic-trends over time, which we consider in this research as means to discover unusual semantic trends of the Incel comment in various times. Overall, we used LDA based on Gibbs sampling to calculate the distance between Incel-comments  and LDA-topics. However, the main aim of the task is to obtain the trends based on semantic aspects of the semantic structure of the online forums in Incel online community. Therefore, by investigating the distributions of these semantic-topics in various days, we obtain semantic trends, as showed in figure 1-2.

\subsection{Strengths and limitations} 
To date, there are limited studies on Incel-comments online analysis to understand the highlight problems and discovering semantic related-words which can be practical to implement targeted strategies on relevant key research fields like cybercrime [39], sociology [40] and other related studies. This research also presents a practical application of AI models based on topic modeling to semantic knowledge discovery in the Incel forums to more understand the characters, problems of Incel issues. For example, according to Table 2-3, we can see ten top most significant topics in more discussion among Incel users based on their discussions of Incel.co website. For example, Topic 83 sounds considerably more generic and is consistent with 'Relationship issues' in general, and marked by \textit{women}, \textit{ugly}, \textit{Incel}, \textit{fuck}, \textit{fucking}, \textit{sex}, \textit{shit}, \textit{guy}, \textit{life}, \textit{men}, \textit{woman}, \textit{face}. Some of the words are very significant to this topic, including  \textit{women}, \textit{ugly}, \textit{Incel}, \textit{men}, \textit{sex}. In addition, some words also cannot have a bright role for labeling this topic. However, the present research has some limitations that should be examined for future. 

\section{Conclusion}
 This paper discusses the new challenge of semantic knowledge discovery and information retrieval in Incel comments of online community. However, a NLP model based on topic modeling for finding the characters and highlighted issues of Incels is utilized in this paper. Moreover, we claim that our result, can be effective to more understand of the behaviors the Incel community based on  the semantics aspects in online forums. We can also mention the role of the Internet and online spaces for investigating, the causes of extremism and, countering radicalization.


%
%



\end{document}